\documentclass[runningheads]{llncs}

\usepackage[T1]{fontenc}
\usepackage{graphicx,verbatim}
\usepackage{amsmath,amsfonts}
\usepackage{booktabs}    % For \toprule, \midrule, \bottomrule
\usepackage{amsmath}     % For math formatting like \pm
\usepackage{graphicx}    % (Optional) In case you include plots
\usepackage{caption}     % (Optional) For customizing captions
\usepackage{pifont}
\usepackage{multirow}  % for multi-line headers
\usepackage{url} % For \url{} in bibliography
\newcommand{\etal}{\textit{et~al.}}

\DeclareUnicodeCharacter{2009}{\,}
\begin{document}
%
% \title{Echo-Tab: Predicting Treatment Strategies for Acute Coronary Syndrome Patients from Echocardiography and Clinical Data}
\title{TREAT-Net: Tabular-Referenced Echocardiography Analysis for Acute Coronary Syndrome Treatment Prediction}
\titlerunning{TREAT-Net: Tabular-Referenced Echocardiography Analysis}
%
\begin{comment}  
%% Removed for anonymized MICCAI 2025 submission
\author{First Author\inst{1}\orcidID{0000-1111-2222-3333} \and
Second Author\inst{2,3}\orcidID{1111-2222-3333-4444} \and
Third Author\inst{3}\orcidID{2222--3333-4444-5555}}

% % First names are abbreviated in the running head.
% % If there are more than two authors, '\etal' is used.
% %
% \institute{Princeton University, Princeton NJ 08544, USA \and
% Springer Heidelberg, Tiergartenstr. 17, 69121 Heidelberg, Germany
% \email{lncs@springer.com}\\
% \url{http://www.springer.com/gp/computer-science/lncs} \and
% ABC Institute, Rupert-Karls-University Heidelberg, Heidelberg, Germany\\
% \email{\{abc,lncs\}@uni-heidelberg.de}}

\end{comment}

\author{Diane Kim\inst{1} \and
Minh Nguyen Nhat To\inst{1} \and
Sherif Abdalla\inst{2,3} \and
Teresa S.M. Tsang\inst{2,3} \and
Purang Abolmaesumi\inst{1} \and
Christina Luong\inst{2,3}}

% \author{Anonymized Authors}  %% Added for anonymized MICCAI 2025 submission
% \authorrunning{Anonymized Author \etal}
\institute{University of British Columbia, Vancouver, BC, Canada \and
Vancouver Coastal Health, Vancouver, BC, Canada  \and
Vancouver General Hospital, Vancouver, BC, Canada}

\authorrunning{D. Kim \etal}
    
\maketitle              % typeset the header of the contribution
\begin{abstract}

Coronary angiography remains the gold standard for diagnosing Acute Coronary Syndrome (ACS). However, its resource-intensive and invasive nature can expose patients to procedural risks and diagnostic delays, leading to postponed treatment initiation. In this work, we introduce \textbf{TREAT-Net}, a multimodal deep learning framework for ACS treatment prediction that leverages non-invasive modalities, including echocardiography videos and structured clinical records. TREAT-Net integrates tabular-guided cross-attention to enhance video interpretation, along with a late fusion mechanism to align predictions across modalities. Trained on a dataset of over $9{,}000$ ACS cases, the model outperforms unimodal and non-fused baselines, achieving a balanced accuracy of 67.6\% and an AUROC of 71.1\%. Cross-modality agreement analysis demonstrates 88.6\% accuracy for intervention prediction. These findings highlight TREAT-Net’s potential as a non-invasive tool for timely and accurate patient triage, particularly in underserved populations with limited access to coronary angiography.
%, with strong reliability and data efficiency. 

\end{abstract}

% Authors must provide keywords and are not allowed to remove this Keyword section.

%
%
%
\section{Introduction}

Acute Coronary Syndrome (ACS) --- commonly referred to as a heart attack --- is characterized by the sudden occlusion of one or more coronary arteries, resulting in the interruption of blood flow to the myocardium. As cardiovascular diseases (CVDs) remain the leading cause of mortality worldwide with an estimated 17.9 million deaths each year \cite{WHO2024}, ACS alone accounts for 23\% and 18\% of these deaths in men and women respectively \cite{timmis2023global}.

Severe cases of ACS require urgent intervention within two hours of onset. In contrast, management strategies for patients with mild to moderate ACS are dependent on the location and severity of occlusion and presenting symptoms \cite{ACSguide}. While non-invasive diagnostic tools such as electrocardiography (ECG) and blood biomarkers can identify certain cardiac abnormalities, they remain insufficient for precise localization and quantification of stenosis \cite{ACSlimit1,ACSlimit2}.  Definitive diagnosis typically relies on invasive coronary angiography --- the current gold standard for visualizing coronary anatomy. Findings from this procedure guide treatment decisions, which may include medical management, percutaneous coronary intervention (PCI), or coronary artery bypass grafting (CABG) \cite{ACSguide}. However, the invasive nature of coronary angiography introduces potential risks including vascular perforation, arterial dissection, acute kidney injury, stroke, and infection \cite{cathrisks}. Moreover, access to angiography is often restricted to tertiary care centers, resulting in the geographic displacement of rural and remote patients \cite{joshi2008global}. These limitations contribute to delays in critical diagnostic and therapeutic decision-making, as well as increased healthcare costs and system-level burden. To eliminate unnecessary angiography in patients suitable for medical management and improve access for those requiring urgent intervention, more effective clinical decision-support tools are required.

% In British Columbia, Canada, only five catheterization laboratories serve a population of approximately 5.7 million, performing over 21,000 diagnostic angiographies annually (570–4). As these facilities are concentrated in a single region, patients residing in rural and remote communities often need to travel significant distances to undergo the procedure. 

Recent studies have demonstrated the potential of deep learning models to advance the analysis and interpretation of medical images \cite{cureus}, enhancing the clinical utility of non-invasive and widely accessible modalities such as ultrasound. These models can offer healthcare providers with rapid and automated assessments of complex patients, particularly in settings with limited access to trained specialists or advanced diagnostic infrastructure.

Among these modalities, echocardiography (echo) is a real-time, non-invasive ultrasound imaging of the heart that provides valuable insights into cardiac structure and function~\cite{echo}.
% \cite{lang2015recommendations,echo}. 
It is rapid, cost-effective, and free from ionizing radiation, making it an attractive alternative to coronary angiography as a front-line diagnostic tool for ACS~\cite{armstrong2010feigenbaum}. Echo enables the detection of regional wall motion abnormalities, assessment of  ventricular function, and identification of complications such as cardiomegaly, all of which are clinically relevant for ACS~\cite{neskovic2013emergency}. Furthermore, echo is highly accessible, as it can be performed at the bedside and is available in smaller community hospitals through the use of point-of-care ultrasound (POCUS) devices. Despite its advantages, however, echo has traditionally not been used as a standalone tool for guiding ACS treatment decisions due to its limited sensitivity and operator dependency~\cite{neskovic2013emergency}.

In this work, we propose \textbf{TREAT-Net} (\textbf{T}abular-\textbf{R}eferenced \textbf{E}cho \textbf{A}nalysis for ACS \textbf{T}reatment), a multimodal deep learning framework that integrates echo videos with routinely collected structured clinical records to classify ACS patients as either medically manageable or requiring active intervention (PCI, CABG). Our contributions are threefold: \textbf{(1)} we develop a scalable multimodal framework tailored for real-world clinical settings where structured electronic medical records (EMR) data and ultrasound devices are readily available but access to angiography is limited; \textbf{(2)} we introduce a modular fusion architecture in which tabular inputs guide cross-attention over imaging embeddings --- providing patient-specific contextualization of echo findings --- followed by late fusion of modality-specific predictions; and \textbf{(3)} we demonstrate that our approach outperforms unimodal and non-fused baselines, enabling more comprehensive, non-invasive, and timely decision-making in ACS care.

\section{Literature Review}

\subsubsection{Deep Learning in Echocardiography.}

Numerous studies have investigated deep learning approaches for echo interpretation, focusing on attention mechanisms to mimic that of expert cardiologists and identify the most clinically relevant features. For example, the Supervised Attention Multiple Instance Learning (SAMIL)~\cite{samil} leverages two attention modules, each to identify clinically relevant views and selectively focus on the most informative images within those views for classification of aortic stenosis (AS) severity. Ahmadi \etal~\cite{tda} introduced frame-level temporal attention mechanisms --- the temporal deformable attention and temporal coherent loss --- to capture not only spatial features but also cardiac motion across frames, which is essential for classifying AS severity. Holste \etal~\cite{holste} proposed EchoCLR that leverages self-supervised learning to pretrain on a larger set of unlabeled data and later finetune on smaller labeled datasets for downstream CVD classification. 

However, many existing studies have limited clinical applicability, as they rely on still images rather than full echo videos, or use either a single video or view per sample. Additionally, they often omit complementary clinical information such as text-based reports or EMR, which are routinely integrated by physicians in real-world decision-making. Such limitations have led to the development of multimodal foundation models that can process full echo studies, integrate complementary clinical information, and generalize to downstream tasks without extensive labeled data. Among these, EchoPrime~\cite{EchoPrime} is a view-informed vision-language model for textual report generation from echo video sequences. Trained on over 12 million samples, its modular design includes separate video and text encoders, a view classifier, and an anatomical attention module that enables task-adaptive multiple instance learning across multi-view inputs. While effective in general cardiac imaging, however, EchoPrime remains underexplored in ACS-specific interpretation or treatment planning. % ~\cite{mil}

% %Such limitations have led to a growing need for multimodal foundation models that can process full echo studies, incorporate various types of modalities, and enable transfer learning to a wide range of downstream tasks without a large set of additional labeled data.

% Among these efforts, EchoPrime~\cite{EchoPrime} is a view-informed vision-language foundation model designed for textual report generation from echo video sequences. Its modular architecture includes video and text encoders, a view classifier, and an anatomical attention module that enables attention-based multiple instance learning~\cite{mil} across multi-view inputs. This design supports flexible and task-adaptive weighting of video tokens based on anatomical structures and clinical targets~\cite{EchoPrime}. While such models have shown success in general cardiac imaging, few have been tested for ACS-specific interpretation or treatment pathway optimization.

\subsubsection{Deep Learning in Structured Tabular Data.}

Until recently, traditional machine learning models --- specifically gradient-boosted decision trees such as XGBoost~\cite{XGBoost} and CatBoost~\cite{CatBoost} 
% and LightGBM~\cite{LightGBM}
--- have continued to show state-of-the-art performance on a wide range of tabular datasets, providing strong baselines with relatively low computational cost. However, the recent emergence of tabular foundation models is rapidly shifting the paradigm for tabular analysis. Tabular Prior-data Fitted Network (TabPFN)v2~\cite{hollmann2025accurate} introduced a transformer~\cite{transformer}-based foundation model trained on millions of synthetic datasets, using in-context learning to jointly process labeled training and unlabeled test samples within a single forward pass. The model treats input-label pairs from the training set as context tokens to approximate the posterior distribution over the label of the test set, mirroring the next-token prediction task in large language models.

In the context of CVD prediction, studies have applied deep learning or ensemble techniques to structured clinical data. Mei \etal~\cite{mei2024optimizing} proposed a gated transformer to predict major adverse cardiovascular events from longitudinal records. Ren \etal~\cite{ren2023predicting} applied a balanced random forest model trained on % SHAP \cite{shap}-
ranked features to forecast acute heart failure following ACS. Alkhamis \etal~\cite{alkhamis2024interpretable} used gradient boosting to capture relationships between cardiac indicators and short-term ACS outcomes. Despite these successes, tabular models alone are limited in their ability to capture dynamic spatial and temporal features of cardiac function, which are especially crucial for borderline or ambiguous ACS cases. Moreover, general-purpose tabular foundation models have not yet been optimized for deployment in time-sensitive triage or treatment planning scenarios, particularly in conjunction with other data modalities.

\section{Method}
\label{sec:method}

\begin{figure}[t]
    \centering
    \includegraphics[width=1\linewidth]{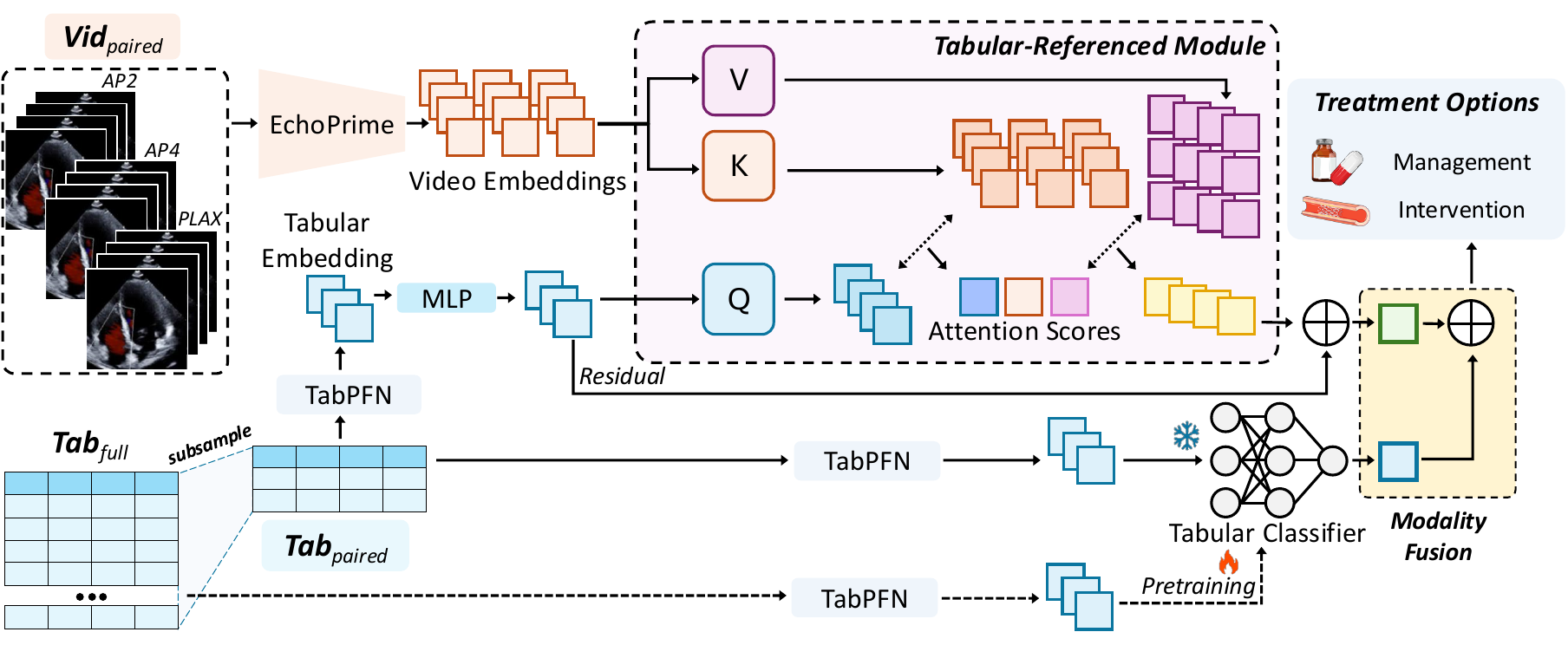}
    \caption{Overview of TREAT-Net architecture.}
    \label{fig:overview}
\end{figure}

\subsection{Framework Overview} 

Figure~\ref{fig:overview} illustrates TREAT-Net's architecture. Given tabular input $x^{\text{tab}} \in \mathbb{R}^{37}$ representing structured clinical data and, for a subset of patients, raw echo videos $x^{\text{vid}} \in \mathbb{R}^{L \times T \times H \times W}$ spanning $L$ echo views of $T$ grayscale frames, the model aims to learn a predictive function that maps $(x^{\text{tab}}, x^{\text{vid}})$ to a binary treatment label. % $y \in \{0, 1\}$.

The training data consists of two sources: a tabular-only dataset $\mathcal{D}_{\text{full}} = \{(x_i^{\text{tab}}, y_i)\}$ and a smaller subset, $\mathcal{D}_{\text{paired}} = \{(x_i^{\text{tab}}, x_i^{\text{vid}}, y_i)\}$, where both structured tabular features and echo videos are available (see Section~\ref{subsection:preprocessing}). 

For effective use of the limited paired data, TREAT-Net first pretrains a tabular classifier on $\mathcal{D}_{\text{full}}$. This is then frozen and reused for downstream multimodal training on $\mathcal{D}_{\text{paired}}$ with echo videos to enhance treatment prediction.

\subsection{Modality-Specific Embedding Extraction}

For each input sample, we obtain two modality-specific embeddings in a shared latent space of dimension \( d = 512 \).

\paragraph{Tabular embedding}
The tabular input $x_{\text{tab}}$ is first encoded using the TabPFN encoder $g_{\text{tab}}$:
\begin{equation}
    \tilde{\mathbf{h}}^{\text{tab}} = g_{\text{tab}}(x^{\text{tab}}) \in \mathbb{R}^{192},
\end{equation}
then subsequently projected onto the shared latent space:
\begin{equation}
    \mathbf{h}^{\text{tab}} = \mathbf{W}_{\text{proj}} \tilde{\mathbf{h}}^{\text{tab}} + \mathbf{b}_{\text{proj}}, \quad \mathbf{W}_{\text{proj}} \in \mathbb{R}^{d \times 192}, \ \mathbf{h}^{\text{tab}} \in \mathbb{R}^d.
\end{equation}

\paragraph{Video embedding}  
The raw video input $x^{\text{vid}}$ is processed through a frozen EchoPrime encoder, which produces a set of $L$ video-level embeddings:
\begin{equation}
    \mathbf{H}^{\text{vid}} = [\mathbf{h}_1^{\text{vid}}, \dots, \mathbf{h}_L^{\text{vid}}] \in \mathbb{R}^{L \times d}.
\end{equation}
Each $\mathbf{h}_\ell^{\text{vid}}$ is a latent representation of the $\ell$-th echo video of a study.

\subsection{Cross-Modality Attention}

We apply multi-head attention (MHA) for clinical features to selectively attend to relevant video representations. The tabular embedding serves as the query $\mathbf{Q} = \mathbf{h}^{\text{tab}} \in \mathbb{R}^{1 \times d}$, while the video sequence provides keys and values $\mathbf{K} = \mathbf{V} = \mathbf{H}^{\text{vid}} \in \mathbb{R}^{L \times d}$. MHA is defined as:
\begin{align}
    \text{head}_i &= \mathrm{softmax}\left( \frac{\mathbf{Q} \mathbf{W}_i^Q (\mathbf{K} \mathbf{W}_i^K)^\top}{\sqrt{d_h}} \right) \mathbf{V} \mathbf{W}_i^V, \\
    \mathrm{MHA}(\mathbf{Q}, \mathbf{K}, \mathbf{V}) &= \mathrm{Concat}(\text{head}_1, \dots, \text{head}_h)\, \mathbf{W}^O,
\end{align}
where $\mathbf{W}_i^Q, \mathbf{W}_i^K, \mathbf{W}_i^V \in \mathbb{R}^{d \times d_h}$, $\mathbf{W}^O \in \mathbb{R}^{d \times d}$, and $d_h = d / h$ is the dimensionality of each head. The output of MHA is denoted $\mathbf{h}^{\text{att}} \in \mathbb{R}^{1 \times d}$.

We compute the fused embedding using residual addition, layer normalization, and a feed-forward network (FFN):
\begin{equation}
    \mathbf{h}^{\text{fused}} = \mathrm{FFN}\left( \mathrm{LayerNorm}(\mathbf{h}^{\text{att}} + \mathbf{h}^{\text{tab}}) \right).
\end{equation}

\subsection{Late Fusion of Modality-Specific Predictions}

% We use two classification pathways: one applied to the fused representation and another derived from the pretrained tabular classifier. 

We use two classification pathways, each using $\tilde{\mathbf{h}}^{\text{tab}}$ and $\mathbf{h}^{\text{fused}}$. Let
\begin{align}
    z^{\text{tab}} &= f_{\text{tab}}(\tilde{\mathbf{h}}^{\text{tab}}), \\
    z^{\text{fused}} &= \mathbf{w}_1^\top \mathbf{h}^{\text{fused}} + b_1, 
    % z^{\text{tab}} &= f_{\text{tab}}(\tilde{\mathbf{h}}^{\text{tab}}),
\end{align}
where $f_{\text{tab}}$ is trained independently on tabular embeddings from $\mathcal{D}_{\text{full}}$ and frozen during multimodal training. The two logits are normalized and averaged:
\begin{equation}
    z^{\text{final}} = \frac{1}{2} \left( \mathrm{LayerNorm}(z^{\text{fused}}) + \mathrm{LayerNorm}(z^{\text{tab}}) \right),
\end{equation}
and the final prediction is:
\begin{equation}
    \hat{y} = \sigma(z^{\text{final}}) = \frac{1}{1 + e^{-z^{\text{final}}}}.
\end{equation}

\section{Experiments}

\subsection{Datasets and Preprocessing}
\label{subsection:preprocessing}

\subsubsection{Tabular Data.} The tabular dataset $\mathcal{D}_{\text{full}}$ contains EMR data of $9{,}906$ ACS patients admitted to a tertiary medical center between 2009 and 2019. Features in $x_i^{\text{tab}}$ include basic demographic information (e.g., age, sex, body mass index), comorbidities (e.g., diabetes, hypertension, hyperlipidemia, smoking, peripheral vascular/cerebrovascular/pulmonary disease, malignancy, heart failure, prior infarction), diagnostic information (e.g., ST-elevated myocardial infarction, non-ST-elevated myocardial infarction, unstable angina), and left ventricular ejection fraction. Numerical columns were imputed using RobustScaler, and categorical columns were processed by OneHotEncoder and OrdinalEncoder. The binary target label $y_i \in \{0, 1\}$ indicates whether the patient received medical management ($y_i = 0$) or active intervention ($y_i = 1$). Class distribution of $\mathcal{D}_{\text{full}}$ is $22\%$ medical management and $78\%$ intervention.

\subsubsection{Echocardiography Videos.}  
% A subset of $\mathcal{D}_{\text{full}}$ forms the paired multimodal dataset $\mathcal{D}_{\text{paired}}$, consisting of $1,571$ patients with a corresponding echo study. For each sample, we extracted the apical-2-chamber (AP2), apical-4-chamber (AP4), and parasternal long-axis (PLAX) videos only, as they are the most clinically relevant for ACS. $16$ frames were extracted with a stride of $2$, which were normalized and resized to $224 \times 224$ pixels. The class distribution for $\mathcal{D}_{\text{paired}}$ is $18\%$ medical management and $82\%$ intervention.
A subset of $\mathcal{D}_{\text{full}}$ forms the paired multimodal dataset $\mathcal{D}_{\text{paired}}$, where $1{,}571$ patients have a corresponding echo study with their tabular records. Across all studies, we extracted $6{,}175$ apical-2-chamber (AP2), $9{,}161$ apical-4-chamber (AP4), and $9{,}752$ parasternal long-axis (PLAX) views, resulting in a total of $25{,}088$ videos. $16$ frames were sampled with a stride of $2$, then normalized and resized to $224 \times 224$ pixels. The class distribution of $\mathcal{D}_{\text{paired}}$ is $18\%$ medical management and $82\%$ intervention.

\subsubsection{Data Splits.} 
% The datasets are randomly split into $70\%$ training, $10\%$ validation, and $20\%$ test set, with class distributions preserved across splits. The split of $\mathcal{D}_{\text{paired}}$ approximately follows the stratified partitioning of $\mathcal{D}_{\text{full}}$, maintaining the same label proportions.
The datasets are randomly split into $70\%$ training, $10\%$ validation, and $20\%$ test sets, with patient exclusivity preserved across splits. The split of $\mathcal{D}_{\text{paired}}$ approximately follows the class-stratified partitioning of $\mathcal{D}_{\text{full}}$.

\subsection{Evaluation}

We compare our proposed model against four baselines: \textbf{(1)} a transformer-based classifier trained on $\mathbf{H}^{\text{vid}}$ from $\mathcal{D}_{\text{paired}}$ without tabular data (EchoPrime+Attn); \textbf{(2)} vanilla TabPFN trained on $\mathcal{D}_{\text{full}}$ (TabPFN); \textbf{(3)} a tabular classifier $f_{\text{tab}}$ trained on $\mathcal{D}_{\text{full}}$ (TabPFN+MLP); and \textbf{(4)} a multimodal model that incorporates tabular and echo data through cross-attention only, without late fusion, relying on a single classification head applied to the fused embedding (Cross-Attention).
% outputs predictions from the video class token
 % vanilla TabPFN trained on $\mathcal{D}_{\text{full}}$ using either TabPFN end-to-end or the frozen TabPFN feature extractor followed by the tabular classifier $f_{\text{tab}}$ described in Section~\ref{sec:method}
 % using either TabPFN end-to-end or the frozen TabPFN feature extractor followed by the tabular classifier $f_{\text{tab}}$ described in Section~\ref{sec:method};

All baseline models are evaluated on the test split of $\mathcal{D}_{\text{paired}}$. Performance is assessed using balanced accuracy (BAcc), Area Under the Receiver Operating Characteristic curve (AUROC), and sensitivity measured at $20\%$, $40\%$, and $60\%$ specificity. Each experiment is repeated five times with unique random seeds.

\subsection{Implementation Details}

All models are implemented in PyTorch and optimized using SGD with a learning rate of $10^{-4}$ and a weight decay of $0.01$. Training is performed with a batch size of $32$ and MHA with $4$ heads. Models are trained for $\approx 40{,}000$ steps using binary cross-entropy loss. 
% The tabular-specific classifier is trained independently on $\mathcal{D}_{\text{full}}$ and frozen when incorporated into TREAT-Net training.

\section{Results}

\begin{table}[t]
\centering
\footnotesize
\caption{Treatment prediction performance. The highest result is \textbf{bolded}, and the second highest result is \underline{underlined}.}

\label{tab:modality_comparison}
\begin{tabular}{lcccccccccc}
\toprule
\multirow{2}{*}{Model} & \multirow{2}{*}{Echo} & \multirow{2}{*}{Tab} & \multirow{2}{*}{Fusion} 
& \multirow{2}{*}{\shortstack{AUROC \\ (\%)}} & \multirow{2}{*}{\shortstack{BAcc \\ (\%)}} 
& \multicolumn{3}{c}{Sensitivity (\%)} \\
\cmidrule(lr){7-9}
& & & & & & 20\% SPE & 40\% SPE & 60\% SPE \\
\midrule
EchoPrime+Attn & \ding{51} & \ding{55} & \ding{55} & 60.2 {\tiny$\pm$ 0.8} & 57.8 {\tiny$\pm$ 1.5} & 87.8 {\tiny$\pm$ 1.0} & 74.7 {\tiny$\pm$ 2.3} & 54.0 {\tiny$\pm$ 4.5} \\
TabPFN & \ding{55} & \ding{51} & \ding{55} & 68.9 {\tiny$\pm$ 0.6} & 63.1 {\tiny$\pm$ 1.1} & \underline{91.2} {\tiny$\pm$ 1.8} & 79.9 {\tiny$\pm$ 1.2} & \underline{67.7} {\tiny$\pm$ 1.8} \\
TabPFN$+$MLP & \ding{55} & \ding{51} & \ding{55} & 68.0 {\tiny$\pm$ 0.1} & 62.6 {\tiny$\pm$ 0.3} & 88.9 {\tiny$\pm$ 0.5} & \underline{80.2} {\tiny$\pm$ 0.6} & 62.9 {\tiny$\pm$ 0.5} \\
Cross-Attention & \ding{51} & \ding{51} & \ding{55} & \underline{69.7} {\tiny$\pm$ 0.6} & \underline{63.4} {\tiny$\pm$ 1.0} & 89.2 {\tiny$\pm$ 2.2} & 78.9 {\tiny$\pm$ 2.4} & 67.1 {\tiny$\pm$ 1.8} \\
\midrule
\textbf{TREAT-Net}  & \ding{51} & \ding{51} & \ding{51} & \textbf{71.1} {\tiny$\pm$ 0.6} & \textbf{67.6} {\tiny$\pm$ 0.5} & \textbf{91.3} {\tiny$\pm$ 1.5} & \textbf{83.3} {\tiny$\pm$ 1.5} & \textbf{70.1} {\tiny$\pm$ 2.7} \\
\bottomrule
\end{tabular}
\end{table}

\subsection{Baselines Comparison}

Table~\ref{tab:modality_comparison} summarizes the treatment prediction performance of the baseline models and TREAT-Net. EchoPrime+Attn shows the lowest performance with an AUROC of $60.2\%$ and BAcc of $57.8\%$. Its intervention prediction sensitivities are the lowest at all specificity thresholds. Tabular data alone through TabPFN and TabPFN+MLP achieves higher numbers across all metrics than video alone, with vanilla TabPFN showing slightly higher performance. Overall, the prediction accuracy using tabular data alone is comparable to the multimodal Cross-Attention approach. However, Cross-Attention shows considerable improvement compared to EchoPrime+Attn, showing as high as $9.5\%$ and $13.1\%$ increase in AUROC and sensitivity at $60\%$ specificity respectively. 

TREAT-Net consistently demonstrates best overall performance over all baselines across all evaluated metrics. Our model obtains an AUROC of $71.1\%$ and a BAcc of $67.6\%$, with increased sensitivities across all specificity thresholds --- up to $16.1\%$ compared to EchoPrime+Attn and $7.2\%$ in TabPFN+MLP. Against Cross-Attention that is similarily multimodal, TREAT-Net still achieves superior performance with up to $4.4\%$ increase in sensitivity at $40\%$ specificity.

\subsection{Ablation Studies}

\subsubsection{Data Efficiency.} 
\begin{figure}[t]
    \centering
    \includegraphics[width=1\linewidth]{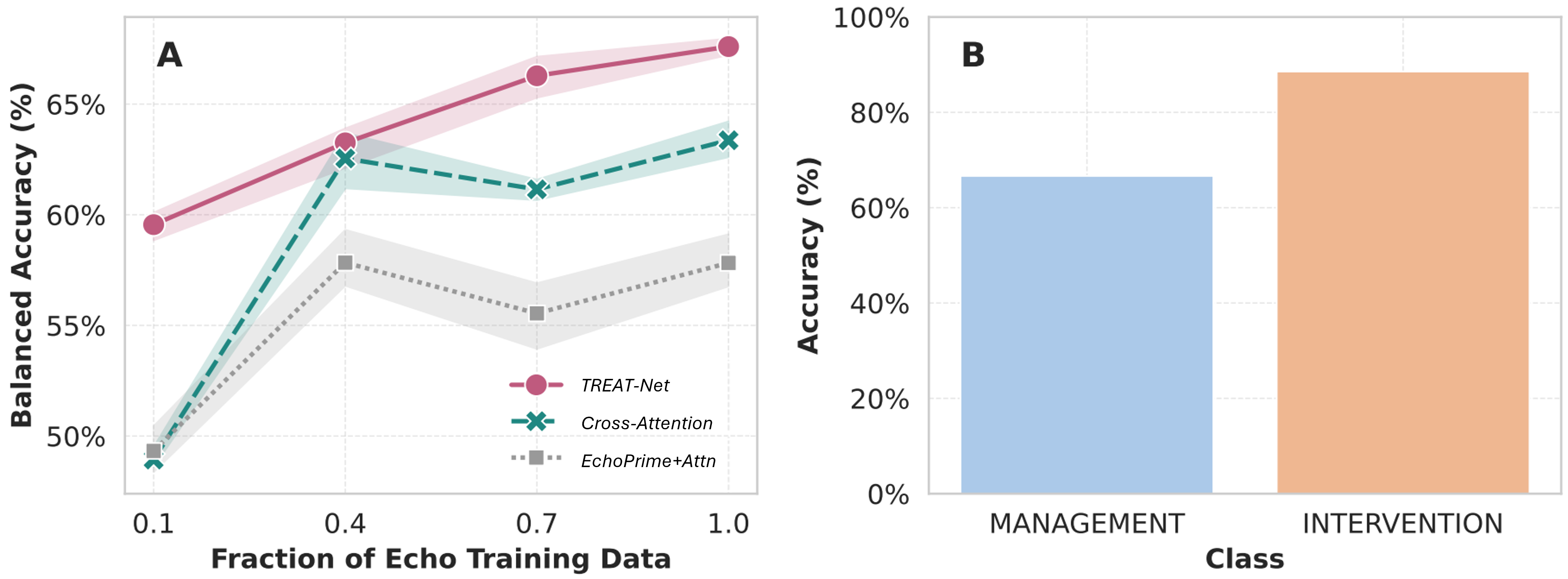}
    \caption{(a) Data efficiency evaluation; (b) Accuracy on agreed predictions.}
    \label{fig:treat_net_ablation}
\end{figure}

Figure~\ref{fig:treat_net_ablation}(a) illustrates how model performance scales with increasing size of $\mathcal{D}_{\text{paired}}$. When trained on $10\%$ of the dataset, both Cross-Attention and EchoPrime+Attn perform below chance level. In contrast, TREAT-Net achieves a BAcc of near $60\%$, benefiting from the pretraining on $\mathcal{D}_{\text{full}}$ and the late fusion architecture. With increasing amounts of echo data, TREAT-Net consistently improves its performance, ultimately achieving a BAcc of $67.6\%$ with the full $\mathcal{D}_{\text{paired}}$, compared to $57.8\%$ for EchoPrime+Attn and $63.4\%$ for Cross-Attention. These results highlight TREAT-Net's strong data efficiency and its capacity for continued performance gain with additional data.
% and its ability to generalize better than either modality alone, especially when more echo imaging data becomes available. 

\subsubsection{Agreement Analysis.} To further assess model reliability, we analyze the subset of test samples where both TREAT-Net and TabPFN+MLP output concordant predictions at $40\%$ specificity, reflecting high-confidence consensus across modalities. Within this subset, class-specific accuracies reach $66.7\%$ for medical management and a striking $88.6\%$ for active intervention. These findings suggest that cross-modal agreement may serve as an implicit confidence signal --- particularly for identifying patients requiring urgent and advanced care. The high precision in intervention classification highlights TREAT-Net's potential use in selective triage and resource prioritization in clinical settings.

\section{Conclusion and Future Directions}

The results of our proof-of-concept study demonstrate the potential for video-tabular multimodal models to distinguish medically manageable ACS patients from those requiring intervention, which is often a clinically challenging task. Leveraging only three standard echo views and routinely collected clinical data, this framework enables early risk stratification at the point of initial presentation. In doing so, it may streamline early decision-making, expedite treatment initiation, reduce the dependency on invasive diagnostic procedures and their associated risks, and optimize the allocation of limited healthcare resources. In particular, our approach can benefit rural and remote settings where access to cardiology expertise is limited and advanced care often involves long wait times and significant geographic displacement from patients' home communities. 
% from patients’ home communities. 

% Particularly, our approach may enable primary care providers to more effectively triage patients to tertiary care centers, particularly in rural and remote settings where 
% For clinically ambiguous cases --- such as patients with borderline (mild to moderate) ACS --- our model may serve as a complementary decision-support tool to aid physicians in making informed, patient-centered judgments aligned with goals of care. 

The next steps for TREAT-Net include incorporating explainability by identifying key tabular features driving treatment predictions and generating saliency maps that highlight echo regions linked to relevant cardiac dysfunctions. Causal effect can also be explored to examine the shift in treatment prediction following changes in specific features, such as age or prior history of infarction. As this study is based on data from a single institution, future research should also evaluate TREAT-Net's scalability and generalizability across diverse geographic regions, healthcare systems, and patient populations. 

\begin{credits}
\subsubsection{\ackname} 

This work was supported in part by the Canadian Institutes of Health Research (CIHR), the Natural Sciences and Engineering Research Council of Canada (NSERC), the Canadian Heart Function Alliance (CHFA),  
% This research was also supposed in part through the computational resources and services provided by 
and the Advanced Research Computing at the University of British Columbia~\cite{UBC_SOCKEYE}. Data for this research was provided by the Provincial Health Services Authority (PHSA). The authors are solely responsible for all inferences, opinions, and conclusions, which may not reflect those of PHSA.

% Data used for this research was provided by the Provincial Health Services Authority (PHSA). All inferences, opinions, and conclusions drawn from this study are those of the authors, and may not reflect the opinions and PHSA's policies.

\subsubsection{\discintname}

The authors have no competing interests to declare that are relevant to the content of this article.

\end{credits}

\bibliographystyle{splncs04}
\bibliography{main}

\end{document}